\documentclass{article}
\usepackage{microtype}
\usepackage{graphicx}
\usepackage{subfigure}
\usepackage{booktabs}
\usepackage{amsmath}
\usepackage{tikz}
\usepackage{xurl}
\usepackage{hyperref}

\usepackage[accepted]{mlsys2025}
\makeatletter
\renewcommand{\Notice@String}{}
\makeatother
\hypersetup{
  hypertexnames=false,
  pdfsubject={Research preprint},
  pdfauthor={Jeongmin Brian Park, Wen-mei Hwu}
}
\mlsystitlerunning{AttentionIndex: Accelerating Learned Sparse Attention}

\begin{document}
\twocolumn[
\mlsystitle{AttentionIndex: Accelerating Learned Sparse Attention\texorpdfstring{\\}{ }with a Causal Inverted-File Index}
\begin{mlsysauthorlist}
\mlsysauthor{Jeongmin Brian Park}{nv}
\mlsysauthor{Wen-mei Hwu}{nv}
\end{mlsysauthorlist}
\mlsysaffiliation{nv}{NVIDIA}
\begin{center}
\small\textsuperscript{1}NVIDIA
\end{center}
\mlsyskeywords{Sparse Attention, Long-Context Inference, Approximate Nearest Neighbor Search, LLM Serving}
\vskip 0.15in
\begin{abstract}
Learned sparse attention, used in production models such as DeepSeek-V4
and GLM-5, lets each query attend to only $k$ keys chosen by a lightweight
indexer, but the indexer still scores every key in the prefix. At one
million tokens, this selection takes more than half of prefill time.
Two-stage selection over contiguous blocks, as in HiSA, reduces this cost,
but its recall declines as context grows at a fixed candidate budget.
We present AttentionIndex, which routes each query over an inverted-file
index that groups indexer keys by similarity and refines the candidates
with the native indexer score. AttentionIndex builds the index during
prefill, making new keys searchable immediately, reusing previous
clustering to keep maintenance cheap, and exposing each query only to
clusters built from its own prefix. On GLM-5.3 and DeepSeek-V4-Flash at
one million tokens, AttentionIndex delivers 1.3--1.8$\times$ the prefill
throughput of exhaustive selection, and at throughput within 5\% of HiSA,
it recovers 17.0 and 10.0 percentage points more of the exhaustive
top-$k$, respectively. AttentionIndex also improves accuracy on long-context benchmarks 
by up to 18.4 percentage points over HiSA at the same candidate budget.
\end{abstract}
]

\section{Introduction}
\label{sec:intro}

Long-context inference has become a defining workload for large language
models: agents~\cite{sun2025contextfolding}, code assistants~\cite{rando2025longcodebench}, and document analysis~\cite{bai2024longbenchv2,zhang2024infinitebench} routinely process
hundreds of thousands to millions of tokens per request~\cite{gemini2024gemini15}. At these lengths,
dense attention is prohibitively expensive, and production models have
adopted learned sparse attention. DeepSeek Sparse Attention
(DSA)~\cite{deepseek2025v32}, used by DeepSeek-V3.2, DeepSeek-V4, and
GLM-5.3~\cite{deepseek2026v4,zai2026glm53config}, adds a lightweight
learned indexer to the model's sparse attention layers. The indexer scores
the keys in the prefix, and the main attention operates only on the $k$
highest-scoring entries, which bounds its per-query cost regardless of
context length.

However, the indexer itself still scores every key. As a result, the computation cost of selection still grows 
linearly with context length per query while
the cost of the attention stays constant. At one million tokens, indexer
scoring and top-$k$ selection take 53--58\% of prefill time on
DeepSeek-V4-Flash and GLM-5.3, more than the rest of the model combined,
even with key compression and cross-layer index reuse
(Section~\ref{sec:indexer-overhead}). At long context, choosing which keys
to attend to has become the bottleneck of sparse attention.

A natural remedy is two-stage selection: route each query to a small set of
candidate keys, then apply the native indexer only to those candidates.
HiSA~\cite{xu2026hisa} routes to contiguous blocks of keys ranked by their
mean-pooled representatives. However, keys of contiguous token entries in the context need not be similar
in indexer space or equally relevant to a query, so a block holding a few
important keys can rank low and be discarded. As context length grows, the probability of entries relevant to a query falling into unselected blocks grows.
In our LongBench-v2
measurements on GLM-5.3, HiSA's recall with 8K candidate keys falls from
83.9\% to 41.8\% as context grows from 16K to 256K tokens. We observe a
similar decline on DeepSeek-V4-Flash with a budget of 4K compressed
entries (Section~\ref{sec:block-routing}).

 We observe that the native indexer performs exhaustive search within each selected block, so
any approximation lies in which keys reside in the selected blocks.
Grouping keys
by similarity rather than by their positions in the context allows the 
exhaustive top-$k$ search to be applied to
fewer groups that contain a higher number of promising candidates, thus reducing the selection process with limited time budget to reach higher quality top-k outcome.
This is the  basic idea behind inverted-file (IVF) indexes for approximate
nearest-neighbor search~\cite{jegou2011pq,johnson2019faiss}. 

 However, three constraints in a learned indexer environment make application of IVF  
more challenging.
First,  since prefill and decode generate new tokens that must be added into the candidate groups, %
updating the IVF index to include these new %
 tokens must cost less than the
scoring work it saves. 
Second,  in order to maintain inference causality, every query must see an IVF index built only from its own
prefix, including the centroids that guide routing. Third, since the indexer
score, i.e. similarity measure, is not generated as a simple inner product, the score for a centroid that is the mean of the members of a cluster is a potentially poor approximate
proxy for how the indexer scores its members.

We present AttentionIndex, which applies the IVF principle with three novel ideas.
First, it routes each query over IVF clusters of %
keys and refines the candidates with the native indexer score, leaving
the sparse attention kernel unchanged. Second, it incrementally builds
and maintains the index during prefill to limit maintenance overhead.
New keys become available to subsequent queries immediately and are
clustered in small batches. Periodic refits regroup the prefix while
reusing the previous clustering and selectively reassessing key
assignments. To preserve causality, each query uses only clusters and
representatives formed from keys in its own prefix. Third, it adjusts
the cluster representatives  to better predict the native indexer score for its cluster members using an encoder
trained offline and shared across the clusters of each layer.

We implement AttentionIndex in SGLang~\cite{zheng2023sglang} and
evaluate it on GLM-5.3 and DeepSeek-V4-Flash with contexts of up
to one million tokens. At one million tokens and almost matched prefill throughput,
AttentionIndex improves recall over HiSA by 17.0 and 10.0 percentage
points on GLM-5.3 and DeepSeek-V4-Flash, respectively. Across the evaluated budgets, AttentionIndex
achieves 1.3--1.8$\times$ the prefill throughput of exhaustive
selection at the same context length. AttentionIndex also improves accuracy on long-context benchmarks by up to 18.4 percentage points over HiSA at the same candidate budget.

In summary, this paper makes the following contributions:
\begin{itemize}
\item We show that in production learned sparse attention, key selection
  rather than attention dominates long-context prefill, and that
  HiSA's recall declines as context grows at a
  fixed candidate budget (Section~\ref{sec:motivation}).
\item We design a causal IVF index for learned indexers that is built
  and maintained during prefill through immediate insertion and
  incremental refits (Section~\ref{sec:design}).
\item We evaluate AttentionIndex on two production models with up to one
  million tokens, measuring selection recall, prefill throughput, and task
  accuracy against exhaustive selection and HiSA
  (Section~\ref{sec:evaluation}).
\end{itemize}

\section{Background}
\label{sec:background}

\subsection{Sparse Attention}

LLM inference processes the input tokens in a prefill phase and then
generates output tokens one at a time during decoding. Dense causal
attention compares each query with every key in its prefix, so its computation cost
grows quadratically with input length during prefill and linearly with
context length at each decoding step. Sparse attention reduces this cost
by restricting each query to a subset of
key--value (KV) entries. Many sparse designs retain a few initial
tokens, which act as attention sinks that receive high attention
regardless of their content~\cite{xiao2023streamingllm}.

The efficiency of sparse attention depends on  the cost of both selecting these
entries and computing attention over them. Systems such as
SampleAttention and LServe use structured sparsity to reduce attention
cost~\cite{zhu2025sampleattention,yang2025lserve}. Grouping entries into
blocks can simplify selection and allow kernels to skip entire blocks.
 However, the fact that a group is not skipped over does not mean that all its entries should be selected for attention computation.
Thus the selection process should consist of two stages: a coarse search can identify candidate groups
before a fine stage selects individual tokens. We call the coarse stage
\emph{routing}, the fine stage \emph{refinement}, and the maximum number
of keys passed from routing to refinement the \emph{candidate budget}.

\subsection{Learned Attention Indexers}

DeepSeek Sparse Attention (DSA) uses a lightweight learned indexer to
select the top-$k$ KV entries for each query~\cite{deepseek2025v32}. The
indexer has its own query and key projections, separate from those of the
main attention. For a query token $t$ and an earlier token $s$, it
computes
\begin{equation}
  I_{t,s} = \sum_{j=1}^{H^I} w^I_{t,j}\,
  \mathrm{ReLU}\!\left(\mathbf{q}^I_{t,j} \cdot \mathbf{k}^I_{s}\right),
  \label{eq:indexer-score}
\end{equation}
where the indexer queries $\mathbf{q}^I_{t,j}$ and weights $w^I_{t,j}$ of
the $H^I$ indexer heads are derived from token $t$, and each token
contributes a single indexer key $\mathbf{k}^I_s$ shared by all heads.
Models that compress keys apply the same score to compressed entries.
The main attention then operates only on the $k$ entries with the highest
scores.

We call $I_{t,s}$ the \emph{native indexer score}. Scoring every available
key with it and keeping the top $k$, as DSA does, is \emph{exhaustive
selection}; it is the reference that faster selection methods approximate
and the Full DSA baseline in our evaluation. We measure an approximate
method by its \emph{recall}: the fraction of the exhaustive top-$k$ that
it recovers.

\subsection{Inverted-File Indexing}

Inverted-file indexing (IVF) speeds up approximate nearest-neighbor
search by restricting candidate evaluation to selected clusters. It
partitions vectors into clusters, commonly using $k$-means, and stores
their members in inverted lists~\cite{jegou2011pq,johnson2019faiss}.
At query time, IVF ranks cluster representatives, selects a subset of clusters,
and evaluates the vectors within those %
clusters to return the top-$k$
candidates. The number of %
 clusters selected, %
or equivalently the number of
 candidate vectors evaluated, controls the tradeoff between search cost and
recall.

Cluster representatives determine which clusters are selected and thus which vectors reach the final scoring
stage. Even when candidates are scored exactly, a true top-$k$ result can
be missed if its cluster is not selected. Standard IVF represents each
cluster by its mean, which suits inner-product search: a query's score
against the mean equals the average score of the cluster's members. The
native indexer score in Equation~\eqref{eq:indexer-score} is not an inner
product, so this equivalence does not hold. IVF also supports insertion by
assigning new vectors to existing clusters without reclustering, and its
clustering cost is typically paid once and amortized over many queries.

\section{Motivation}
\label{sec:motivation}

\subsection{The Cost of Exhaustive Indexer Scoring}
\label{sec:indexer-overhead}

Learned sparse attention bounds the cost of the main attention computation,
but not the cost of choosing which entries it attends to. In DSA, attention
uses only the top-$k$ entries for each query, yet exhaustive selection
scores every available indexer key and selects the $k$ highest
scores~\cite{deepseek2025v32}. For a context of $N$ tokens, scoring and
top-$k$ selection each require $O(N)$ work per query and $O(N^2)$ work
across prefill, assuming fixed model dimensions and compression ratios,
whereas attention over the selected entries requires $O(k)$ work per query.
The cost of key selection therefore grows with context length, while the
per-query cost of sparse attention stays constant.
This holds in both phases of
inference: selection is quadratic across a prefill, and every decoding
step must again score the entire prefix.

Recent models reduce this overhead through compression and reuse.
DeepSeek-V4-Flash compresses every four tokens into one indexed entry in
its Compressed Sparse Attention (CSA) layers~\cite{deepseek2026v4}, and
GLM-5.3 reuses indexer selections across multiple attention
layers~\cite{zai2026glm53config}, exploiting the high overlap between the
selections of adjacent layers~\cite{bai2026indexcache}. These optimizations
reduce the number of keys or the frequency of scoring, but each fresh
selection still scans the available indexed prefix.

\begin{figure}[h]
  \centering
  \includegraphics[width=\columnwidth]{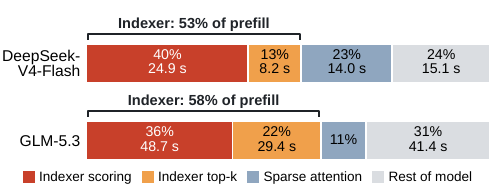}
  \caption{Prefill time breakdown for one 1M-token request under
  exhaustive selection.  ``Rest of
  model'' includes MoE and dense GEMMs, communication, normalization, RoPE,
  quantization, compressor kernels, and idle time.}
  \label{fig:prefill-breakdown}
\end{figure}

Figure~\ref{fig:prefill-breakdown} shows that these optimizations do not
remove the bottleneck at long context. In a 1M-token prefill, indexer
scoring and top-$k$ selection together take 53\% of prefill time on
DeepSeek-V4-Flash and 58\% on GLM-5.3, more than the rest of each model
combined and 2.4$\times$ and 5.2$\times$ the time spent in sparse
attention, respectively. On GLM-5.3, top-$k$ selection alone takes nearly
twice as long as sparse attention, so an effective remedy must shrink the
input to top-$k$ as well as the number of keys scored.

AttentionIndex targets this selection cost by routing each query to a
bounded set of candidate keys before applying the native indexer, so both
scoring and top-$k$ selection operate on these candidates rather than the
full prefix. The objective is to recover the keys chosen by exhaustive
selection with less computation, while keeping inference-time index
construction and maintenance below the savings they provide.

\subsection{Limitations of Contiguous-Block Routing}
\label{sec:block-routing}

Block-level routing reduces selection cost by scoring one representative
for many keys. MoBA~\cite{lu2025moba} ranks contiguous blocks using
mean-pooled keys, and HiSA~\cite{xu2026hisa} applies the same idea to the
learned indexer. It selects blocks using their summaries, then refines
the keys in those blocks with the native indexer score. A fixed candidate
budget bounds refinement work, but routing determines which keys
refinement can recover.

\begin{table}[h]
\caption{HiSA recall (\%) of the exhaustive top-$k$ at fixed candidate
budgets: 8K token keys for GLM-5.3 and 4K C4 entries for DeepSeek-V4-Flash.
Context lengths count source tokens; dashes indicate that the budget
covers the context.}
\label{table:hisa-recall}
\vskip 0.15in
\begin{center}
\begin{small}
\setlength{\tabcolsep}{4pt}
\begin{tabular}{@{}lrrrr@{}}
\toprule
 & \multicolumn{2}{c}{GLM-5.3} & \multicolumn{2}{c}{DeepSeek-V4-Flash} \\
 & \multicolumn{2}{c}{HiSA8K} & \multicolumn{2}{c}{HiSA4K} \\
\cmidrule(lr){2-3}\cmidrule(lr){4-5}
Context & LongBench-v2 & RULER & LongBench-v2 & RULER \\
\midrule
16K  & 83.9 & 81.7 & --   & --   \\
32K  & 73.4 & 68.3 & 87.4 & 82.0 \\
48K  & 66.5 & 60.0 & 83.7 & 70.4 \\
64K  & 59.3 & 55.9 & 80.3 & 64.7 \\
96K  & 53.1 & 50.5 & 65.7 & 57.2 \\
128K & 46.6 & 48.2 & 59.3 & 50.6 \\
256K & 41.8 & 41.4 & 46.0 & 41.8 \\
\bottomrule
\end{tabular}
\end{small}
\end{center}
\vskip -0.1in
\end{table}

Table~\ref{table:hisa-recall} shows that HiSA's recall drops as context
grows at a fixed candidate budget. On GLM-5.3, recall falls from 83.9\%
to 41.8\% on LongBench-v2 and from 81.7\% to 41.4\% on RULER as context
grows from 16K to 256K tokens. On DeepSeek-V4-Flash, it falls from 87.4\%
to 46.0\% on LongBench-v2 and from 82.0\% to 41.8\% on RULER between
32K and 256K tokens. At 256K, HiSA misses more than half of the exhaustive
top-$k$ in all four settings. 

At a fixed block size and candidate budget, longer contexts introduce
more blocks competing for the same routing slots. The challenge is to
represent a block's useful keys with a single centroid. Contiguous keys
need not be similar in indexer space or equally relevant to a query, so
their mean can score poorly even when a few members belong to the
exhaustive top-$k$. Once their block is discarded, refinement cannot
recover those keys. HiSA's failure analysis identifies this loss from
mean pooling~\cite{xu2026hisa}. 

This motivates grouping keys by their indexer embeddings rather than
their positions. AttentionIndex clusters similar keys, allowing routing
to retrieve related keys together without also admitting  less relevant  their positional
neighbors.

\section{Design}
\label{sec:design}

\begin{figure*}[t]
  \centering
  \IfFileExists{Figures/design_overview.pdf}{
    \includegraphics[width=\textwidth]{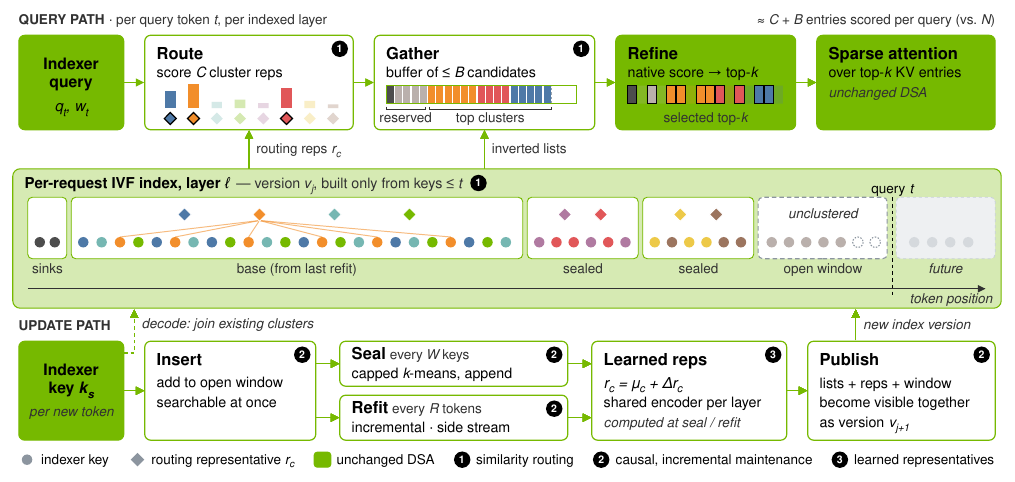}
  }{
    \fbox{\parbox[c][1.4in][c]{0.95\textwidth}{\centering
      Overview figure goes here (Figures/design\_overview.pdf)}}
  }
  \caption{Overview of AttentionIndex for one indexed layer. \textbf{Query
  path:} routing scores cluster representatives with the native indexer
  score and gathers the keys of the top-ranked clusters into the candidate
  budget, after reserved slots for sink and window keys; refinement selects
  the top-$k$ candidates for sparse attention. \textbf{Update path:} new
  keys enter an open window and are searchable at once; every $W$ keys the
  window is sealed into local clusters, and every $R$ tokens a refit
  regroups the prefix. A new index version is visible only to queries whose
  prefix contains all of its keys.}
  \label{fig:design-overview}
\end{figure*}

AttentionIndex replaces exhaustive selection with routing over an
inverted-file (IVF) index of indexer keys, followed by refinement with the
native indexer score. Because the index grows with the requests that use
it, during both prefill and decoding, two requirements shape the design:
maintaining the index must cost less than the scoring work that it saves, and every
query must see an index built only from its own prefix. The design has three parts
(Figure~\ref{fig:design-overview}): routing over similarity clusters
(Section~\ref{sec:overview}), causal and incremental index maintenance
(Sections~\ref{sec:insertion} and~\ref{sec:refit}), and learned routing
representatives (Section~\ref{sec:learned-centroids}).

\subsection{Routing over Similarity Clusters}
\label{sec:overview}

For each request and indexed layer, AttentionIndex keeps an IVF index over
the indexer keys~\cite{jegou2011pq,johnson2019faiss}. Each cluster stores
an inverted list of its member key IDs and a routing representative. A
query scores every representative with the native indexer score
(Equation~\eqref{eq:indexer-score}) and ranks the clusters. It then fills a
candidate budget of $B$ keys: sink keys and the not-yet-clustered keys of
the open window take reserved slots, and the lists of the top-ranked
clusters fill the rest, with the last list truncated if needed. Refinement
scores these candidates with the native indexer score and passes the
top-$k$ keys to the unchanged sparse attention kernel.

Like HiSA, AttentionIndex selects keys in two stages, but its routing units
are clusters of similar keys rather than contiguous blocks. Because
refinement scores every candidate exactly, all approximation comes from
candidate selection: a key in the exhaustive top-$k$ is missed only if it
never reaches the candidate set. Grouping keys by similarity concentrates
the exhaustive top-$k$ in fewer clusters, so the same budget covers more of
it (Section~\ref{sec:retrieval-quality}).

Each query scores about $C+B$ entries instead of $N$, where $C$ is the
number of clusters. With about 128 keys per cluster on GLM-5.3, routing
still grows linearly with the prefix but with a far smaller constant, and
refinement is bounded by $B$.

\subsection{Causal Maintenance of a Growing Index}
\label{sec:insertion}

The prefix grows with every token, so the index must change while queries
use it. Rebuilding it whenever new keys arrive would erase the savings,
while only adding keys to existing clusters lets clusters grow unevenly.
AttentionIndex therefore separates immediate insertion from lazy
reorganization.

New keys enter an open window and are included directly in the candidates
of later queries, so they are searchable before they are clustered. Every
$W$ keys, AttentionIndex \emph{seals} the window: it clusters the window's
keys with size-capped $k$-means and appends the new clusters without
modifying published ones. On GLM-5.3, a seal groups $W=1{,}024$ keys into
eight clusters of about 128 keys. Until the first seal, queries use
exhaustive selection.

Sealing keeps insertion cheap but confines each cluster to one window, so
similar keys that arrive in different windows stay apart. Every $R$ tokens
($R=16{,}384$ on GLM-5.3), a \emph{refit} therefore regroups the entire
available prefix into a new base, with more clusters as the prefix grows.
The window size trades candidate slots for insertion work; the refit
interval sets how long window boundaries persist. During decoding, which
adds one key per step, new keys instead join existing clusters directly,
because a typical response stays below the refit interval.

Causality must hold for routing, not only for candidate IDs: a
representative computed from a future key could change which clusters a
query selects even if that key is later masked. Each seal or refit
therefore produces a new index version whose lists, representatives, and
window metadata become visible together, and a query uses a version only if
every key that built it lies in the query's prefix. Within a prefill chunk,
earlier queries keep the previous version and cover newer keys through
their own window.

\subsection{Incremental Refits}
\label{sec:refit}

Two properties let a refit reuse the previous base instead of clustering
from scratch: consecutive refits share almost all of their keys, and
indexer keys never change once generated. AttentionIndex warm-starts each
refit from the previous base's nonempty means plus new centers sampled from
recently added keys. It also keeps, for every key, its assignment, an upper
bound on the distance to its center, and lower bounds on the distances to
competing centers~\cite{elkan2003,drake2012adaptive}. A refit compares all
keys with the new centers, but only newly added keys and keys whose bounds
no longer certify their assignment search all centers. With $\Delta C$ new
centers and $M$ keys that need a full search, the distance work is
proportional to $N\Delta C+MC$ instead of $NC$. Each refit runs a single
assignment and update pass, trading exact convergence for lower cost.

A refit executes on a separate
stream while the forward pass continues, and each layer publishes its new
version before the first query that needs it. The new base replaces the
clusters of the prefix it covers and keeps any seals made since; only
sealing remains on the forward path.

\subsection{Learned Routing Representatives}
\label{sec:learned-centroids}

Cluster membership and routing serve different purposes. Membership uses
geometric distance, which is cheap to update and supports the bounds above.
Routing ranks clusters by the native indexer score, under which a cluster's
mean is only an approximate proxy for its members
(Section~\ref{sec:background}), so scoring the mean can rank a cluster low
even when it contains top-$k$ keys. AttentionIndex therefore keeps
membership geometric and learns only the representatives used for routing,
following \citet{vecchiato2024learning}.

Because clusters are created and replaced within each request, a separate
representative cannot be optimized per cluster. Instead, a small encoder
shared by all clusters of a layer predicts a correction $\Delta r_c$ to the
cluster mean $\mu_c$ from permutation-invariant summaries of its members:
their mean, the logarithm of the cluster size, and mean- and max-pooled
learned features~\cite{zaheer2017deepsets}. The routing representative is
$r_c=\mu_c+\Delta r_c$. The correction is bounded and initialized to zero,
so the encoder starts from ordinary IVF and leaves membership, means, and
bounds unchanged.

We train the encoder offline on queries and index snapshots from the frozen
model, with exhaustive selection as the reference. For a training query,
let $h_c$ count the reference keys in cluster $c$ and $n_c$ its causally
visible members, both excluding mandatory candidates, and let $z_c$ be the
native indexer score of $r_c$. We minimize the cross entropy between
\begin{equation}
  t_c=\frac{h_c}{\sum_j h_j}
  \quad\text{and}\quad
  p_c=\frac{n_c\exp(\alpha z_c)}{\sum_j n_j\exp(\alpha z_j)},
  \label{eq:routing-loss}
\end{equation}
plus a penalty on $\Delta r_c$, where $\alpha$ is a training scale. With
unconstrained scores the optimum satisfies
$\exp(\alpha z_c)\propto h_c/n_c$, so the encoder learns to rank clusters
by reference keys per candidate slot. The encoder runs only when a seal or
refit creates clusters; queries score representatives exactly as in
ordinary IVF.

\section{Implementation}
\label{sec:implementation}

We integrate HiSA and AttentionIndex into SGLang as candidate producers
for the learned indexer~\cite{zheng2023sglang}. Full DSA and both selectors
share the serving stack, model weights, sparse attention kernels, and
client. 
The selectors use Triton and TileLang kernels and read paged indexer keys
directly. On GLM, routing representatives use the same FP8 page format as
indexer keys, allowing routing and refinement to reuse DeepGEMM.

Refits read quantized keys directly and run on separate CUDA streams,
with work staggered across layers to overlap the forward pass. Under
tensor parallelism, indexer keys are replicated, while each layer has an
owner rank that fits and broadcasts its index state so all ranks use the
same routing geometry. Learned encoders are trained offline and loaded
at server startup. Layers
without a selected encoder use ordinary cluster means.

We measure recall by capturing indexer operands during serving and
replaying selectors under the serving schedule. Replay checks require
HiSA to match its served selections and IVF to preserve the server's
mandatory candidates. Serving timings include index construction and
maintenance, but exclude startup and kernel compilation. Time breakdowns
use PyTorch profiler traces of a prefill. Accuracy runs use greedy decoding
and each benchmark's answer extraction and metric, applied after the
model's thinking segment.

\section{Evaluation}
\label{sec:evaluation}

\subsection{Experimental Setup}
\textbf{Models.} We evaluate AttentionIndex on GLM-5.3 and
DeepSeek-V4-Flash. We preserve their native selection sizes:
2,048 token keys per query for GLM-5.3 and 512 compressed entries
for the indexed layers of DeepSeek-V4-Flash, which use a compression
stride of four source tokens.

\textbf{Baselines.} We compare against Full DSA, which exhaustively
scores the causal key set using the native indexer, and
HiSA~\cite{xu2026hisa}, which selects contiguous blocks using pooled
key summaries before scoring their member keys.

\textbf{Datasets.} We use LongBench-v2~\cite{bai2024longbenchv2},
RULER~\cite{hsieh2024ruler}, and
InfiniteBench~\cite{zhang2024infinitebench}, covering code repositories,
long-document question answering, in-context learning, and synthetic
retrieval tasks.

\textbf{Hardware.} Timing experiments use eight NVIDIA B200 GPUs with
eight-way tensor parallelism and 8,192-token prefill chunks. All methods
run in the same serving stack (Section~\ref{sec:implementation}).

\subsection{Indexer Retrieval Quality Analysis}
\label{sec:retrieval-quality}

In this section, we measure recall as the fraction of the exhaustive
native indexer's top-$k$ keys recovered in the final selection.
All methods and budgets use the same
12 inputs per dataset and model. LongBench-v2 uses ten code-repository and two in-context
learning inputs, excluding data used to train or select learned
representatives and taking prefixes of approximately 1M tokens. RULER
uses two 512K inputs from each of six tasks, and InfiniteBench uses 12 long contexts with at most one per
book. We evaluate the final 2,048 queries of each prefix at five indexed
layers spanning the model's depth, averaging over queries and layers
within each input, then across inputs. Budgets span 16K--64K candidates
($1\mathrm{K}=1{,}024$), counted as token keys on GLM and compressed C4
entries on V4. The V4 budgets therefore correspond to 64K--256K
source-token equivalents.

\begin{figure*}[t]
  \centering
  \includegraphics[width=\textwidth]{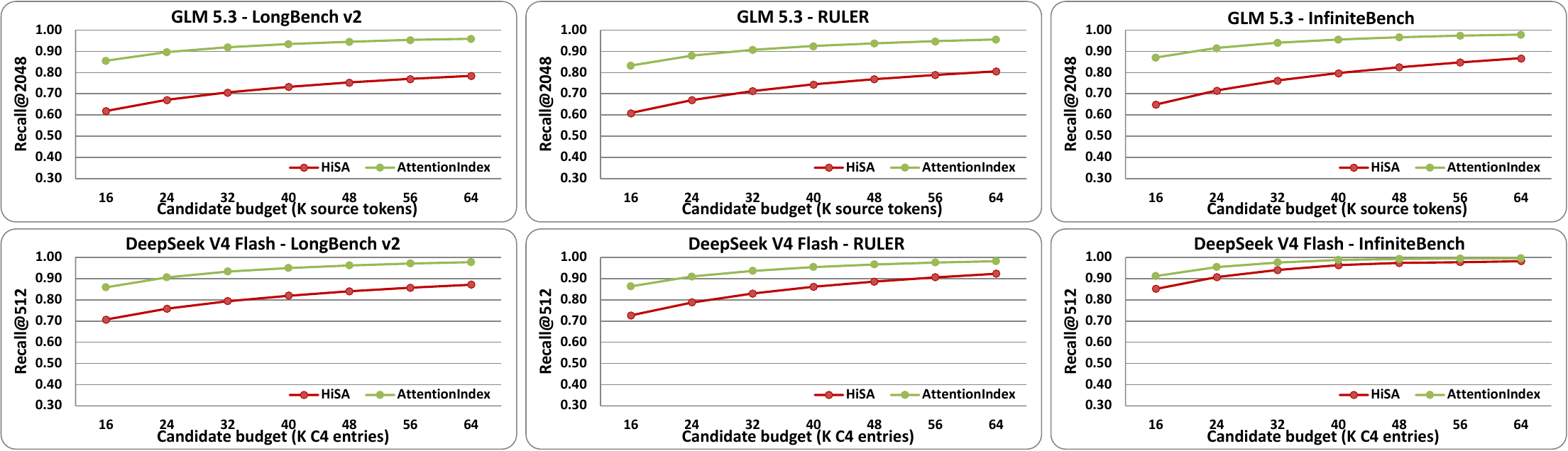}
  \caption{Indexer selection recall of HiSA and AttentionIndex on
  GLM-5.3 and DeepSeek-V4-Flash across three datasets.
  Candidate budgets count token keys for GLM and compressed C4 entries
  for DeepSeek-V4-Flash.}
  \label{fig:retrieval-quailiy}
\end{figure*}

AttentionIndex achieves higher mean recall than HiSA (Figure~\ref{fig:retrieval-quailiy}) on all three
datasets at every evaluated budget supported by both methods.
For example, at a 24K budget on LongBench-v2, AttentionIndex improves
recall from 67.16\% to 89.70\% on GLM-5.3 and from 75.88\% to 90.66\%
on DeepSeek-V4-Flash.
The smaller gains on DeepSeek-V4-Flash at larger budgets reflect the higher recall
already achieved by HiSA. 

AttentionIndex also achieves higher recall with smaller candidate
budgets. On GLM-5.3, its 24K results exceed HiSA's 64K results on all
three datasets, using 62.5\% fewer candidate slots. On DeepSeek-V4-Flash, its 24K-entry
results exceed HiSA's 32K-entry results on all three datasets, using
25\% fewer slots. Increasing the budget further improves recall for
both methods. These results
demonstrate that AttentionIndex uses the candidate budget more
effectively, although the resulting latency benefit also depends
on routing and index maintenance costs.

The gains are consistent with the limitations of HiSA's contiguous-block
means~\cite{xu2026hisa} (Section~\ref{sec:block-routing}): similarity
grouping preserves more reference keys within the same candidate budget.

\subsection{Throughput Analysis}
\label{sec:throughput}

A single-stream client evaluates prefix cuts of long
LongBench-v2 documents, with two repeats per point and one server launch
per method. We report effective prefill throughput as
$N/\mathrm{TTFT}$ in tokens per second, where $N$ is the number of input
tokens and TTFT is the median time from request submission to the first
generated token. Timing includes index construction and maintenance
throughout prefill.

\begin{figure*}[t]
  \centering
  \IfFileExists{Figures/Eval-Throughput.pdf}{%
    \includegraphics[width=\textwidth]{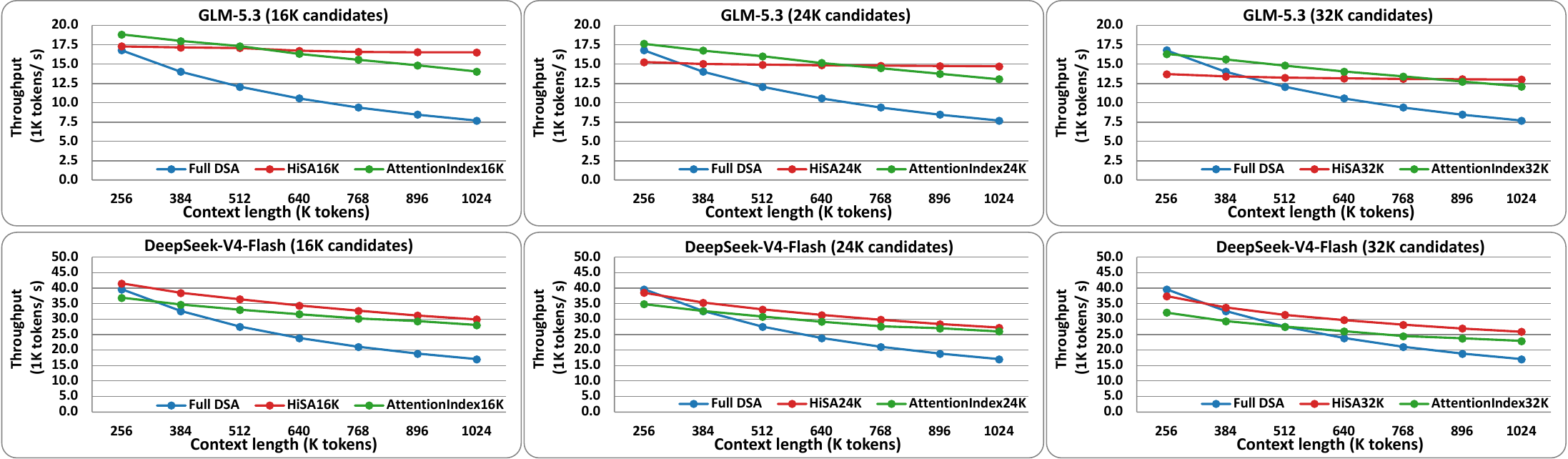}%
  }{%
    \fbox{\parbox[c][1.4in][c]{0.95\textwidth}{\centering
      Throughput figure goes here (Figures/Eval-Throughput.pdf)}}%
  }
  \caption{Effective prefill throughput ($N/\mathrm{TTFT}$) as context
  length grows, for Full DSA and for HiSA and AttentionIndex at 16K, 24K,
  and 32K candidate budgets on GLM-5.3 and DeepSeek-V4-Flash.}
  \label{fig:throughput}
\end{figure*}

\textbf{Scaling with context length.} Full DSA loses more than half of its
throughput between 256K and 1M tokens on both models (Figure~\ref{fig:throughput}), because every query
scores the entire prefix. Both approximate methods decline far less. At 1M
tokens, AttentionIndex reaches 1.6--1.8$\times$ the throughput of Full DSA
on GLM-5.3 and 1.3--1.6$\times$ on DeepSeek-V4-Flash across the three
budgets, and its advantage grows with context length. At 256K, where
exhaustive selection is still cheap, the index has not yet paid for itself
at every budget; AttentionIndex matches or exceeds Full DSA at all budgets
from 384K tokens on GLM-5.3 and from 512K on DeepSeek-V4-Flash.

\textbf{Comparison at equal budgets.} At the same budget,
AttentionIndex is faster than HiSA on GLM-5.3 up to 512K--768K tokens,
depending on the budget, but slower at longer contexts and on
DeepSeek-V4-Flash. Unlike HiSA, AttentionIndex maintains its index and
gathers candidates from scattered positions rather than contiguous blocks,
and maintenance grows with the prefix because each refit revisits all
available keys. This is consistent with its throughput declining with
context length while HiSA's stays nearly constant on GLM-5.3. However,
AttentionIndex achieves higher recall at the same budget
(Section~\ref{sec:retrieval-quality}), so throughput alone does not capture
the difference between the methods.

\textbf{Recall at comparable throughput.}
Table~\ref{table:recall-throughput} pairs AttentionIndex16K with HiSA
configurations whose measured throughput differs by less than 5\% at
512K and 1M tokens. Throughput is normalized to AttentionIndex16K
separately for each model and context length. Recall is measured on four
LongBench-v2 code-repository inputs, averaging five indexed layers and
the final 2,048 body query positions of each prefix.

\begin{table}[t]
\caption{Recall (\%) at comparable prefill throughput. Tput is throughput
relative to AttentionIndex16K for each model and context. Context lengths
count source tokens; budgets count token keys for GLM and compressed C4
entries for DeepSeek-V4-Flash. Dashes denote exhaustive selection.}
\label{table:recall-throughput}
\centering
\small
\setlength{\tabcolsep}{1.5pt}
\begin{tabular}{@{}lrrr@{\hspace{6pt}}rrr@{}}
\toprule
 & \multicolumn{3}{c}{\textit{GLM-5.3}} & \multicolumn{3}{c}{\textit{DeepSeek-V4-Flash}} \\
\cmidrule(lr){2-4}\cmidrule(l){5-7}
Method & Budget & Tput & Recall & Budget & Tput & Recall \\
\midrule
\multicolumn{7}{l}{\textit{512K tokens}} \\
Full DSA       & --  & 0.698 & 100.0 & --  & 0.833 & 100.0 \\
HiSA           & 16K & 0.987 &  63.1 & 24K & 1.002 &  85.2 \\
AttentionIndex & 16K & 1.000 &  86.4 & 16K & 1.000 &  89.5 \\
\midrule
\multicolumn{7}{l}{\textit{1M tokens}} \\
Full DSA       & --  & 0.548 & 100.0 & --  & 0.608 & 100.0 \\
HiSA           & 24K & 1.048 &  69.6 & 24K & 0.971 &  76.5 \\
AttentionIndex & 16K & 1.000 &  86.6 & 16K & 1.000 &  86.5 \\
\bottomrule
\end{tabular}
\end{table}

AttentionIndex achieves higher recall in all four comparisons. At 512K
and 1M tokens, it gains 23.3 and 17.0 percentage points on GLM, and 4.3
and 10.0 points on DeepSeek-V4-Flash. These gains hold at comparable speed, including
three comparisons where AttentionIndex uses one third fewer candidates.
Full DSA provides reference recall of 100\%, but achieves only
0.55--0.83 times AttentionIndex16K's throughput at these operating points.

\subsection{Decoding Latency Under Load}
\label{sec:tpot}

HiSA reports larger TPOT gains as request concurrency
increases~\cite{xu2026hisa}. We evaluate decoding latency when a long
prefill request arrives during ongoing generation.
Table~\ref{table:tpot-load} lists the two workloads.
We report mean TPOT over the first 512 output tokens, including
interruptions from the arriving prefill, averaged over two repeats.
The node's 1.1M-token KV capacity limits the concurrency we can
evaluate at these context lengths.

\begin{table}[h]
\caption{TPOT over the first 512 output tokens (ms), with speedup
over Full DSA in parentheses.}
\label{table:tpot-load}
\centering
\small
\setlength{\tabcolsep}{2pt}
\begin{tabular}{@{}lrrr@{}}
\toprule
Decoders & $4\times32$K & $2\times128$K  \\
Arrival  & 917K & 786K \\
\midrule
Full DSA
  & 231.3 (1.00$\times$)
  & 184.1 (1.00$\times$) \\
HiSA24K
  & 140.0 (1.65$\times$)
  & 119.3 (1.54$\times$) \\
AttentionIndex16K
  & 162.7 (1.42$\times$)
  & 127.4 (1.45$\times$) \\
AttentionIndex24K
  & 163.9 (1.41$\times$)
  & 131.6 (1.40$\times$) \\
\bottomrule
\end{tabular}
\end{table}

AttentionIndex achieves TPOT speedups of 1.40--1.45$\times$
over Full DSA, compared with 1.54--1.65$\times$ for HiSA24K.
KV capacity limits these workloads to two or four concurrent
decoders. Larger batches would better amortize routing overhead and decrease
 the gap between HiSA and AttentionIndex.

\subsection{End-to-End Accuracy}
\label{sec:accuracy}

We evaluate task accuracy using 200 InfiniteBench questions
(25 from each of eight tasks), 198 RULER samples (66 each at 128K, 256K,
and 512K), and LongBench-v2. InfiniteBench and RULER use complete
contexts.

\begin{table}[h]
\caption{Task accuracy (\%) with exhaustive selection (Full DSA) and with
HiSA and AttentionIndex at 16K and 24K candidate budgets. Budgets count token
keys for GLM and compressed C4 entries for DeepSeek-V4-Flash.}
\label{table:accuracy}
\vskip 0.15in
\begin{center}
\begin{small}
\setlength{\tabcolsep}{3pt}
\begin{tabular}{lccccc}
\toprule
 & Full & \multicolumn{2}{c}{HiSA} & \multicolumn{2}{c}{AttentionIndex} \\
\cmidrule(lr){3-4}\cmidrule(lr){5-6}
Dataset & DSA & 16K & 24K & 16K & 24K \\
\midrule
\multicolumn{6}{l}{\textit{GLM-5.3}} \\
\quad InfiniteBench & 83.0 & 75.0 & 78.6 & 80.4 & 83.3 \\
\quad RULER         & 81.8 & 55.2 & 62.6 & 73.6 & 77.5 \\
\quad LongBench-v2  & 63.8 & 60.4 & 61.0 & 60.4 & 61.7 \\
\midrule
\multicolumn{6}{l}{\textit{DeepSeek-V4-Flash}} \\
\quad InfiniteBench & 81.16 & 80.46 & 80.19 & 81.20 & 82.03 \\
\quad RULER         & 77.83 & 58.59 & 67.27 & 74.92 & 76.14 \\
\quad LongBench-v2  & 63.82 & 60.82 & 63.02 & 61.63 & 63.42 \\
\bottomrule
\end{tabular}
\end{small}
\end{center}
\vskip -0.1in
\end{table}

Table~\ref{table:accuracy} reports task accuracy on both models, with
Full DSA as the reference. At a matched 24K budget, AttentionIndex
improves accuracy over HiSA on all three benchmarks for both models.
The largest gains are on RULER: 14.9 percentage points on GLM and 8.87
points on DeepSeek-V4-Flash, recovering approximately 78\% and 84\% of HiSA's gap to
Full DSA. On InfiniteBench, the gains are 4.7 and 1.84 points,
respectively. 

The benefit is smaller on LongBench-v2. Full DSA exceeds HiSA24K by
only 2.8 points on GLM and 0.80 points on DeepSeek-V4-Flash, compared with 19.2 and
10.56 points on RULER. AttentionIndex24K recovers 0.7 and 0.40 points,
respectively. The small gap to exhaustive selection suggests limited
accuracy headroom from improving indexer recall alone. LongBench-v2
emphasizes reasoning over long contexts~\cite{bai2024longbenchv2}, so
model capability may remain a limiting factor even when relevant keys
are selected. The aggregate scores support this interpretation but do
not identify the cause of each missed question.

DeepSeek-V4-Flash's smaller InfiniteBench gain is also consistent with its higher
starting recall. At 24K entries, HiSA already recovers 90.74\% of the
exhaustive top-$k$, and AttentionIndex raises recall to 95.51\%
(Section~\ref{sec:retrieval-quality}). On GLM, the corresponding change
is much larger, from 71.58\% to 91.59\%. HiSA's DeepSeek-V4-Flash accuracy is already
close to Full DSA, at 80.19\% versus 81.16\%, and AttentionIndex reaches
82.03\%. These results show that the accuracy benefit depends on both
the retrieval gap and the task; an improvement in indexer recall need
not produce a proportional improvement in accuracy.

Finally, the advantage largely persists at comparable throughput.
AttentionIndex16K and HiSA24K have measured prefill throughput within
5\% on both models at 1M tokens (Section~\ref{sec:throughput}).
Using these configurations, AttentionIndex achieves higher accuracy in
four of the six comparisons while using one third fewer
candidates. It gains 1.8 and 11.0 points on GLM's InfiniteBench and RULER,
and 1.01 and 7.65 points on DeepSeek-V4-Flash's. LongBench-v2 is the exception: its
scores are lower by 0.6 and 1.39 points, respectively.

\begingroup
\setlength{\textfloatsep}{12pt plus 2pt minus 2pt}
\subsection{AttentionIndex Time Breakdown}
\label{sec:time-breakdown}

\begin{figure}[h]
  \centering
  \includegraphics[width=\columnwidth]{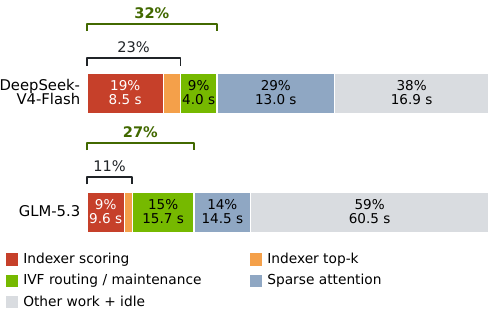}
  \caption{GPU work during AttentionIndex prefill at approximately 1M
  tokens and a 24K budget in token keys (GLM) or C4 entries (DeepSeek-V4-Flash).}
  \label{fig:attnidx-breakdown}
\end{figure}

Figure~\ref{fig:attnidx-breakdown} separates AttentionIndex's routing
and maintenance from the remaining model work. This category includes
candidate packing and remapping, index construction and refitting,
centroid encoding, publication, and index exchange across TP ranks.
We attribute GPU work using operation ranges and maintenance-thread
launches, so background refits and encoder GEMMs are included even when
their kernel names resemble ordinary model operations. Scoring and
top-$k$ kernels, including those used for coarse routing, are counted
separately to avoid duplication.

Scoring and top-$k$ together account for 11.6\,s on GLM and 10.3\,s on
DeepSeek-V4-Flash, or 11\% and 23\% of the totals. Routing and maintenance add
15.7\,s (15\%) and 4.0\,s (9\%), respectively. Including these costs
raises the share of all index work to 27\% on GLM and 32\% on DeepSeek-V4-Flash, as
shown by the longer brackets. Sparse attention accounts for a further
14\% and 29\%, leaving 59\% and 38\% for other work and idle time.

These shares describe GPU work rather than added request latency:
kernel durations on concurrent streams are summed, and background
refits overlap the forward computation. End-to-end throughput and
the incremental cost of learned centroids use separate serving measurements
(Sections~\ref{sec:throughput} and~\ref{sec:ablation-learned}).

\subsection{Ablation Studies}
\label{sec:ablation}

\subsubsection{Periodic Refitting}
\label{sec:ablation-refit}

We compare
AttentionIndex with refitting disabled against several refit intervals
at fixed candidate budgets.

\begin{table}[h]
\caption{Mean indexer recall with different refit intervals and candidate
budgets. Both quantities count token keys for GLM and compressed C4
entries for DeepSeek-V4-Flash; 1K denotes 1,024.}
\label{table:ablation-refit}
\centering
\small
\setlength{\tabcolsep}{3pt}
\renewcommand{\arraystretch}{1.05}
\begin{tabular}{@{}lrr@{\hspace{8pt}}rr@{}}
\toprule
 & \multicolumn{2}{c}{\textit{GLM-5.3}} & \multicolumn{2}{c}{\textit{DeepSeek-V4-Flash}} \\
\cmidrule(lr){2-3}\cmidrule(l){4-5}
Refit interval & 24K & 32K & 24K & 32K \\
\midrule
None (seal only) & 0.788 & 0.825 & 0.8166 & 0.8524 \\
4K              & 0.880 & 0.904 & 0.9131 & 0.9388 \\
8K              & 0.886 & 0.910 & 0.9089 & 0.9358 \\
16K             & 0.883 & 0.907 & 0.8995 & 0.9282 \\
32K             & 0.879 & 0.905 & 0.8849 & 0.9163 \\
\addlinespace[2pt]
HiSA            & 0.665 & 0.702 & 0.7610 & 0.7972 \\
\bottomrule
\end{tabular}
\end{table}

Table~\ref{table:ablation-refit} shows that every tested refit interval
improves recall over sealing alone at every budget. At the serving
interval of 16K, GLM recall rises from 0.788 to 0.883 with 24K candidates
and from 0.825 to 0.907 with 32K candidates, gains of 9.49 and 8.22
percentage points using unrounded measurements. All 13 evaluated GLM
inputs improve at both budgets. On DeepSeek-V4-Flash, the same refit interval adds
8.29 points at 24K and 7.58 points at 32K. Its sweep uses 13 documents:
12 from LongBench-v2 and one from RULER. These results show that
regrouping the growing prefix recovers more reference keys during
candidate selection than sealing alone.

On GLM, most of the benefit comes from performing periodic refits.
Changing the interval to 4K, 8K, or 32K changes mean recall by at most
0.37 points relative to the served 16K setting. The 8K interval gives
the highest mean recall, exceeding 16K by 0.27 and 0.23 points at the two
budgets; shortening it further to 4K does not improve recall. DeepSeek-V4-Flash is more
sensitive to the interval, with shorter intervals consistently yielding
higher mean recall in this sweep. The benefit over sealing alone is
generally larger at smaller candidate budgets, where routing has less room to
compensate for poor grouping.

\subsubsection{Learned Centroids}
\label{sec:ablation-learned}

Table~\ref{table:ablation-learned-recall} compares ordinary $k$-means
centroids and learned representatives with fixed membership and a
24K budget of token keys for GLM or compressed entries for DeepSeek-V4-Flash. GLM's
learned layers use ten LongBench-v2 inputs at 512K, with 8,192 query
positions per input. DeepSeek-V4-Flash uses 36 inputs across the three datasets,
evaluated at each prompt's end.

\begin{table}[t]
\caption{Selected-layer recall with ordinary and learned centroids.
Gains are percentage points, computed before
rounding recall. A dagger marks layers using the ordinary centroid in
both configurations. GLM layer 14 is enabled in this ablation.}
\label{table:ablation-learned-recall}
\centering
\small
\setlength{\tabcolsep}{1.5pt}
\begin{tabular}{@{}lrrr@{\hspace{6pt}}lrrr@{}}
\toprule
\multicolumn{4}{c}{\textit{GLM-5.3}} & \multicolumn{4}{c}{\textit{DeepSeek-V4-Flash}} \\
\cmidrule(r){1-4}\cmidrule(l){5-8}
Layer & $k$-means & Learned & Gain & Layer & $k$-means & Learned & Gain \\
\midrule
0               & 0.928 & 0.962 & $+3.41$ & 2             & 0.885 & 0.888 & $+0.23$ \\
14              & 0.846 & 0.896 & $+5.01$ & 12            & 0.931 & 0.934 & $+0.22$ \\
34              & 0.875 & 0.880 & $+0.55$ & 22$^{\dagger}$ & 0.934 & 0.934 & $0.00$ \\
54              & 0.885 & 0.896 & $+1.12$ & 32$^{\dagger}$ & 0.941 & 0.941 & $0.00$ \\
74$^{\dagger *}$ & 0.895 & 0.895 & $0.00$ & 42$^{\dagger}$ & 0.924 & 0.924 & $0.00$ \\
\bottomrule
\end{tabular}
\par\vspace{3pt}
\parbox{\columnwidth}{\footnotesize
$^{*}$GLM layer 74 reports checkpoint-selection validation
candidate-pool recall, averaged over 24K and 32K budgets.}
\end{table}

The improvement varies across layers: GLM gains 0.55--5.01 percentage
points across the four layers with learned representatives. DeepSeek-V4-Flash gains 0.23 and 0.22
points at layers 2 and 12, with improvements on 32 and 30 of the 36
documents, respectively. 

DeepSeek-V4-Flash's compressed entries mix source tokens through overlapping
windows~\cite{deepseek2026v4}; routing adjustments leave this mixture unchanged.
This may limit the gains, although the ablation does not isolate
compression's effect.

\begin{table}[t]
\caption{TTFT (s) with ordinary and learned centroids in matched serving
runs at a 24K budget in each model's native units. $\Delta$ is the relative
TTFT change, computed before rounding.}
\label{table:ablation-learned-cost}
\centering
\small
\setlength{\tabcolsep}{2pt}
\begin{tabular}{@{}lrrr@{\hspace{7pt}}rrr@{}}
\toprule
 & \multicolumn{3}{c}{\textit{GLM-5.3}} & \multicolumn{3}{c}{\textit{DeepSeek-V4-Flash}} \\
\cmidrule(lr){2-4}\cmidrule(l){5-7}
Context & $k$-means & Learned & $\Delta$ (\%) & $k$-means & Learned & $\Delta$ (\%) \\
\midrule
512K & 31.58 & 31.85 & $+0.9$ & 17.58 & 17.71 & $+0.8$ \\
896K & 65.79 & 65.86 & $+0.1$ & 35.32 & 35.89 & $+1.6$ \\
1M   & 77.03 & 78.19 & $+1.5$ & 41.72 & 41.62 & $-0.2$ \\
\bottomrule
\end{tabular}
\end{table}

Table~\ref{table:ablation-learned-cost} isolates encoder cost with matched
hardware and index settings, changing only whether learned representatives
are enabled. Each context uses one prompt and medians over two to six
repetitions. TTFT changes by $+0.1$\% to $+1.5$\% on GLM and
$-0.2$\% to $+1.6$\% on DeepSeek-V4-Flash; measured prefill throughput changes by less
than 1.6\% in every case. The encoder runs only when clusters are created
or refitted, leaving query scoring unchanged
(Section~\ref{sec:learned-centroids}).

\endgroup

\section{Related Work}
\label{sec:related}

\textbf{Sparse attention.} Structured approaches keep local windows and attention
sinks~\cite{xiao2023streamingllm} or exploit structured sparsity in
attention kernels~\cite{zhu2025sampleattention,yang2025lserve}. Query-aware
methods select pages of the KV cache at inference time~\cite{tang2024quest},
and trainable architectures learn block or token selection together with
the model~\cite{yuan2025nsa,lu2025moba,deepseek2025v32}. Routing
Transformers cluster queries and keys with online $k$-means during
training~\cite{roy2021routing}. AttentionIndex does not change how a model
attends: it accelerates the selection step of a learned sparse attention
mechanism at inference time and preserves its native score.

\textbf{Accelerating learned indexers.} HiSA replaces the exhaustive indexer
scan with routing over contiguous blocks~\cite{xu2026hisa}, the approach we
compare against throughout. IndexCache reuses selections across
layers~\cite{bai2026indexcache}, and kernel-level work streams indexer
scores to bound memory~\cite{jaber2026streamindex} or speeds up exact
top-$k$ selection~\cite{liu2026hpcopstopk}. These optimizations largely
complement AttentionIndex's reduction in the number of keys scored.

\textbf{Retrieval over the KV cache.} RetrievalAttention builds approximate
nearest-neighbor indexes over keys~\cite{liu2024retrievalattention}, and
RetroInfer organizes the KV cache as a vector storage
engine~\cite{chen2026retroinfer}. ClusterKV, Squeezed Attention, and Tactic
cluster keys to select or estimate important
tokens~\cite{liu2024clusterkv,hooper2024squeezed,zhu2026tactic}.
MagicPIG samples keys with locality-sensitive hashing~\cite{chen2025magicpig},
while SAAP partitions keys and queries asymmetrically using a learned query
classifier~\cite{mazare2025saap}. These methods approximate
dense attention, mostly during decoding, with indexes built
offline or after prefill. AttentionIndex instead targets the top-$k$ of a
learned indexer and maintains its index within the forward pass, keeping
routing causal as the prefix grows.

\textbf{Dynamic vector indexes and clustering.} Dynamic indexes combine
incremental insertion with periodic
reorganization~\cite{xu2023spfresh,baranchuk2023dedrift}. AttentionIndex's
seals and refits resemble buffered writes and merges in log-structured
merge trees~\cite{oneil1996lsm}. Its refits build on bound-based $k$-means
acceleration~\cite{elkan2003,hamerly2010,drake2012adaptive,ding2015yinyang},
and its learned representatives follow a learning-to-rank formulation of
cluster selection~\cite{vecchiato2024learning}. We adapt these techniques
to an index that is built within a single request and publishes causal
updates on the inference path.

\section{Conclusion}
\label{sec:conclusion}

Learned sparse attention reduces attention computation, but exhaustive key
selection remains costly at long contexts. AttentionIndex accelerates
selection by maintaining a causal IVF index within each request and
refining routed candidates with the native indexer score.
At a matched 24K candidate budget, AttentionIndex improves recall and task accuracy over HiSA on both models, while delivering 1.3--1.8$\times$ the prefill throughput
of exhaustive selection at one million tokens. Optimizing its per-step
decoding cost remains future work.

\bibliography{reference}
\bibliographystyle{mlsys2025}

\section*{AI Usage Disclosure}
We used Claude Opus, Fable 5.1, GPT-6 Astra, and OpenAI Codex to assist
with code generation, literature search, and writing. For implementation,
AI assistance included generating and revising portions of the code for
the proposed system. For literature search, it helped identify relevant
papers, summarize their methods, and compare them with our approach.
For writing, assistance included developing section outlines,
drafting and reorganizing paragraphs, and revising explanations for
clarity and concision.
\end{document}